\documentclass[10pt,conference]{IEEEtran}

\usepackage[english]{babel}
\usepackage[T1]{fontenc}
\usepackage[utf8]{inputenc}
\usepackage{cite}
\ifCLASSINFOpdf
\usepackage[pdftex]{graphicx}
\graphicspath{{figs/}}
\DeclareGraphicsExtensions{.pdf,.jpeg,.png}
\else
\usepackage[dvips]{graphicx}
\graphicspath{{../figs/}}
\DeclareGraphicsExtensions{.eps}
\fi
\usepackage[cmex10]{amsmath}
\usepackage{amsthm}

\usepackage{algorithmic}
\usepackage{array}
\ifCLASSOPTIONcompsoc
\usepackage[caption=false,font=normalsize,labelfont=sf,textfont=sf]{subfig}
\else
\usepackage[caption=false,font=footnotesize]{subfig}
\fi
\usepackage{float}
\usepackage{units}
\usepackage{xcolor,colortbl}
\usepackage{graphicx}
\usepackage{hyperref}
\usepackage{multirow}       
\usepackage{tikz}
\usetikzlibrary{shapes,arrows,chains}
\tikzset{
	vertex/.style = {
		circle,
		fill            = black,
		outer sep = 2pt,
		inner sep = 1pt,
	}
}

\usepackage[export]{adjustbox}
\usepackage{url}
\begin{document}
	\title{Robust statistics and no-reference image quality assessment in Curvelet domain.}
	
	%------------------------------------------------------------------------- 
	% change the % on next lines to produce the final camera-ready version 
	\newif\iffinal
	%\finalfalse
	\finaltrue
	\newcommand{\jemsid}{99999}
	%------------------------------------------------------------------------- 
	
	
	\iffinal
	
	\author{\IEEEauthorblockN{Ramon Giostri Campos, Evandro Ottoni Teatini Salles}
		\IEEEauthorblockA{Programa de Pós-Graduação em Engenharia Elétrica\\
			Universidade Federal do Espírito Santo\\
			Vitória, ES, 29075-910, Brasil\\
			Email:ramon.campos@ufes.br, evandro.salles@ufes.br }
            %\and
            }
	
	
	%Telephone: (800) 555--1212, Fax: (888) 555--1212}
	
	\else
	\author{SIBGRAPI paper ID: \jemsid \\ }
	\fi
	
	\maketitle
	
\begin{abstract}
\selectlanguage{english} 
This paper uses robust statistics and curvelet transform to learn a general-purpose no-reference (NR) image quality assessment (IQA) model. The new approach, here called M1, competes with the Curvelet Quality Assessment proposed in 2014 (Curvelet2014). The central idea is to use descriptors based on robust statistics to extract features and predict the human opinion about degraded images. To show the consistency of the method the model is tested with 3 different datasets, LIVE IQA, TID2013 and CSIQ. To test evaluation, it is used the Wilcoxon test to verify the statistical significance of results and promote an accurate comparison between new model M1 and Curvelet2014. The results show a gain when robust statistics are used as descriptor.
\end{abstract}
\IEEEpeerreviewmaketitle


%\selectlanguage{brazil} 
\section{Introduction}
\label{s:intro}

\textit{This is a non-doubly blind version of the article accepted for publication in the XIV Workshop de Vis\~{a}o Computacional\footnote{\url{http://www.wvc2018.com.br/proceedings}}. This version differs from that published only because it has links to download all parts of the software developed and has one additional figure. The software, trained models and some useful routines can be downloaded from the link \url{https://github.com/rgiostri/robustcurvelet}}.
	
The Image Quality Assessment (IQA) is an important subject for computational vision, which impacts in robotics, super-resolution, reconstruction of images, evaluation of medical images, among other areas.  The research in IQA aim in developing algorithms to make an objective quality score of images (Q) consistent with human opinion, regardless of the content of the image, the level and type of distortion.

This area has three main groups of methods, Full-Reference (FF),Reduced-Reference (RR)  and No-Reference (NR). In applied sciences there is usually no reference image, therefore NR methods are appropriate. For research areas like astronomy, microscopy, remote sensing and medical images, the NR models are the unique way to evaluate quality of images without consulting an expert. 

This work proposes a NR method, oriented to natural scenes which the  features are extracted by the transformed space of curvelets\cite{Candes2006}. The traditional Wavelets transform do not have orientation, and orientation facilitates the study of anisotropy in degraded images\cite{Gabarda2007}. The parametrization of Curvelets uses position, scale and orientation, which favors if compared to the traditional Wavelet for this study.

%A vantagem mais evidente do uso de Curvelets é sua parametrização incluir orientação, além de posição e escala. As tradicionais Wavelets não possuem orientação, prejudicando a representação de imagens com estruturas orientadas. A orientação facilita o estudo da anisotropia nas imagens degradadas.

The architecture used in this paper was proposed by  Moorthy and Bovik\cite{Moorthy2010} that uses the several modules of Support Vector Machine (SVM) organized in two stages (2-stages SVM).

\begin{figure}[H]
\centering
\includegraphics[width=.5\textwidth,center]{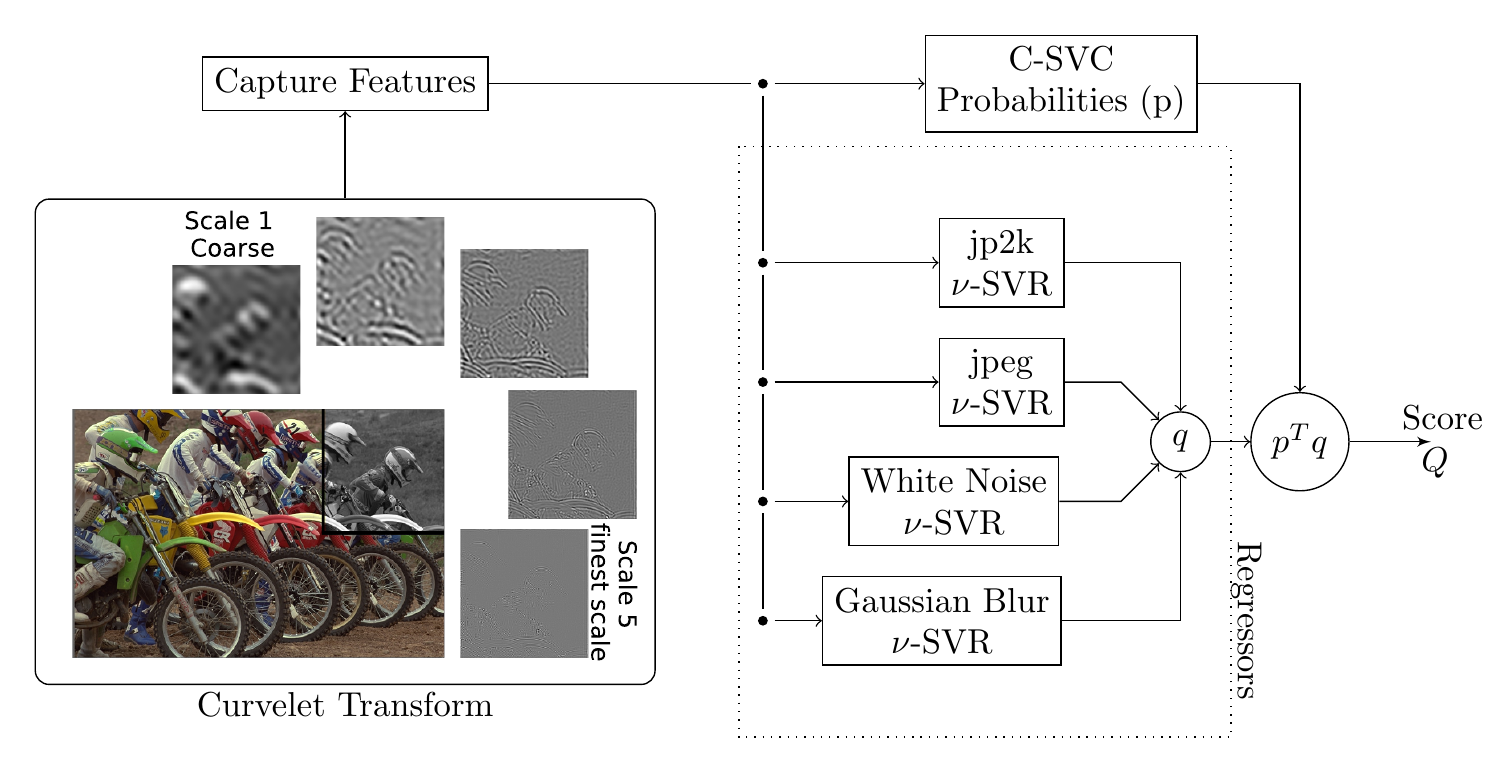}
\caption{Flowchart shows the structure of the 2-stages SVM for 4 degradations.}
\label{fig_sim}
\end{figure}

The complete explanation of 2-stages SVM is in subsection \ref{ss:2stages} and the flow chart can be viewed in Figure  \ref{fig_sim}. 
	
In the stage 1, the classifier C-support vector classification (C-SVC), provides the probabilities of an image to be of a specific class. In the second stage, each regressor nu-support vector regression ($\nu$-SVR), provide her prediction. The fusion of predictions is performed by a weighted average. The weights are given by the probability of C-SVC stage.
	
This work searchs forms to improve the feature extraction in Curvelets space, using robust and non-parametric statistics, which is the main difference for the model proposed by Liu et al.\cite{Liu-2014}, here called Curvelet2014. The two proposals are close by the use of 2-stages SVM, and features extracted in Curvelets space, but are distant by the way the features are extracted.

%Diferentemente do que foi proposto em Liu et al.\cite{Liu-2014}, aqui denominada de Curvelet2014, este trabalho procura mecanismos para melhorar a eficiência da extração de características no espaço das Curvelets, se posicionando como uma alternativa mais eficiente e robusta. As duas propostas se aproximam por usarem a arquitetura 2-\textit{stages} SVM, e extraírem características no espaço das Curvelets, mas diferem pela forma como as características são extraídas.
	
The propose defined in this paper, called model M1, uses orthodox tools of robust statistics based on quartiles, octiles\cite{Groeneveld1984}\cite{Moors1988} and the Median Absolute Deviation (MAD)\cite{Rousseeuw1993}. The M1 uses features to describe images and at same time exempts the analysis of outliers and statistical models.

%A proposta defendida nesse artigo, denominada modelo M1, utiliza de forma ortodoxa ferramentas estatísticas robustas baseada em quartis e octis\cite{Groeneveld1984}\cite{Moors1988} e no desvio absoluto da mediana(MAD)\cite{Rousseeuw1993}. O modelo M1 utiliza características que procuram refletir o tipo de degradação sofrida pelas imagens ao mesmo tempo que exime a análise da influência de valores atípicos (\textit{outliers}).

The competing proposes are trained with the LIVE IQA\cite{liveDatabase}  database and tested with the TID2013\cite{TID2013}, CSIQ\cite{CSIQ} and the LIVE IQA. It is important to mention that because the selected datasets the conclusions are restricted to natural scene images. This type of image is defined as any image obtained on ordinary commercial cameras. The degradation classes used in this paper are compressions jpeg (jpeg) and jpeg2000 (jp2k), Gaussina white noise (wn), Gaussian blur (gblur). One experiment was done using the classes of degradation simultaneously and other experiment for type of degradation. In all the experiments we used the statistical significance test of Wilcoxon\cite{Wilcoxon1945} to make strengthen conclusions.
	
%Os modelos concorrentes são treinados no banco de dados LIVE IQA\cite{liveDatabase} e testados nos bancos de dados TID2013\cite{TID2013} e CSIQ\cite{CSIQ}, além do próprio LIVE IQA. Isso delimita as conclusões ao ramo das imagens de cenas naturais. Esse tipo de imagem é definida como qualquer imagem óticas obtidas em câmeras comerciais ordinárias. As classes de degradação estudadas foram a compressão jpeg (jpeg) e jpeg2000 (jp2k), ruído branco Gaussino (wn), borramento Gaussiano (gblur). Foram feitos experimentos usando as classes de degradação simultaneamente e experimentos por tipo de degradação. Em todos os experimentos se recorreu ao teste de significância estatística de Wilcoxon\cite{Wilcoxon1945} para robustecer as conclusões.
	
All choices made in this article are shown to facilitate their use and reproducibility. An examples of the code can be downloaded in github page \url{https://github.com/rgiostri/robustcurvelet}.

%Todas as escolhas feitas neste artigo são mostradas visando que sua reprodução e que seu uso possa ser facilitado.

The contributions of this article to the scientific community are: 1) Propose a new ways to extract features in the transformed space of Curvelets. These features are based on robust statistical descriptors. 2) Replace the model Curvelet2014, due to the better performance of the new method, which is less dependent on training data and has a better correlation with human perception. The final product is a competitive, open source and reproducible IQA method.

%As contribuições desse artigo para ciência são: 1) propor novas formas de extrair características no espaço transformado das Curvelets. 2) Substituir o modelo de referência Curvelet2014. O produto final é um método IQA competitivo, de uso aberto e reprodutível.

The organization of this work is as follows: In Section \ref{s:RelatedWork} shows a review of the work related to NR IQA, focusing mainly on the use of the Curvelet transform. In Section \ref{s:PropAproach} shows the invocation. In Section \ref{s:method} presents the methodology. In Section \ref{s:exp_res} shows the experiments, results and discussions. Finally, in Section \ref{s:Conclusion} the conclusions of the study are shown.
	
%A organização desse trabalho segue da seguinte maneira: A Seção \ref{s:RelatedWork} mostra uma revisão dos trabalho relacionados a NR IQA, focando principalmente no uso da transformada Curvelet. A Seção \ref{s:PropAproach} trata da inovação proposta. A seção \ref{s:method} apresenta a metodologia de trabalho. A Seção \ref{s:exp_res} trata dos experimentos, resultados e discussões. Na Seção \ref{s:Conclusion} são mostradas as conclusões do trabalho.
	
\section{Related Works}
\label{s:RelatedWork}

This work uses the second generation Curvelets transform proposed by Candes et al.\cite{Candes2006}, called Fast Discret Curvelet Transform (FDCT). The implementation used was the Frequency Wrapping (FDCT-Wrap), avaliable for download on the website of CurveLab\footnote{\url{http://www.curvelet.org/}}. The binding with python used was PyCurvelab,  available here\footnote{\url{https://github.com/slimgroup/PyCurvelab}}. 

The first use of FDCT in NR IQA proposals was made by Shen et al.\cite{Shen2009}, using only the finest layer for feature extraction.
	
%Esse trabalho utiliza a segunda geração de transformadas Curvelets proposta por Candes et al.\cite{Candes2006}, chamada de \textit{Fast Discret Curvelet Transform} (FDCT), a forma de implementação utilizada é a \textit{Frequency Wrapping} (FDCT-Wrap) disponível para download no sítio dos próprios autores. O primeiro uso da FDCT em propostas NR IQA foi feito por  Shen et al.\cite{Shen2009}.
	
The architecture 2-stages SVM was introduced by Moorthy and Bovik\cite{Moorthy2010} applied to features extracted from natural scenes. This architecture was used in by \cite{Zhao2015}\cite{Ahmed2017a}\cite{Mittal2012b}\cite{Moorthy2012} with different features.

%A arquitetura utilizando 2-\textit{stages} SVM foi introduzida por Moorthy e Bovik\cite{Moorthy2010} para características extraídas usando estatísticas de cenas naturais. Essa arquitetura foi empregada nos trabalhos \cite{Zhao2015}\cite{Ahmed2017a}\cite{Mittal2012b}\cite{Moorthy2012} utilizando outras formas de extrair características. 

The use of FDCT with 2-stages SVM was proposed by Liu et al.\cite{Liu-2014}. Zhao et al.\cite{Zhao2015} proposed an NR IQA method that uses the entropy of the Curvelet transform coefficients, also associated with the 2-stages SVM. Ahmed e Der\cite{Ahmed2017a} used the method proposed in \cite{Liu-2014} with the spatial features for study of contrast in images. Recently Shahkolaei et al.\cite{SHAHKOLAEI2018} use the model Curvelet2014\cite{Liu-2014} to quantify the quality of images associated with historical documents.

% O uso da FDCT junto a arquitetura 2-\textit{stages} SVM foi proposto por Liu et al.\cite{Liu-2014}. Os autores Zhao et al.\cite{Zhao2015} propuseram um método NR IQA que utiliza a entropia dos coeficientes da transformada Curvelet, também associada a arquitetura 2-\textit{stages} SVM. Os autores Ahmed e Der\cite{Ahmed2017a} utilizaram o método proposto em \cite{Liu-2014} associado a características espaciais das imagens degradadas para o estudo de contraste em imagens. E mais recentemente, no trabalho de Shahkolaei et al.\cite{SHAHKOLAEI2018}, o modelo Curvelet2014\cite{Liu-2014} foi utilizado para quantificar a qualidade das imagens de documentos históricos.

%**Sair**Outras arquiteturas em dois estágios podem ser encontradas nos trabalhos de \cite{Zeng2017}\cite{Saad2012}\cite{Kang2014}.**
    
Examples of NR IQA work using non-parametric statistics associated with the Curvelet transform can be found in \cite{Shen2009}\cite{Zhao2015}. On the other hand, in the literature review, the authors did not find any work in NR IQA that made use of robust statistical tools to extract features from the Curvelets space.

%Exemplos de trabalhos em NR IQA utilizando estatística não paramétrica associada a transformada Curvelet podem ser encontrados em \cite{Shen2009}\cite{Zhao2015}. Por outro lado, no levantamento bibliográfico realizado, os autores não encontraram qualquer trabalho em NR IQA que fizesse uso de ferramentas estatísticas robustas para extração de características oriundas do espaço das Curvelets.

\section{Proposed Aproach}
\label{s:PropAproach}
	
\subsection{Pre-processing}

Follow \cite{Candes2006}, the total number of Curvelet coefficients and the total number of scales ($n_s$) depends on the image size (M,N) and the number of directions ($\chi$) on second coarse scale ($S_2$). This work uses 32 directions on $S_2$  and all images are fragmented in blocks with size (256,256) pixels. The others scales are: $S_1$ is the coarse layer with only one orientation, $S_3$ with 64 orientations, $S_4$ the fine scale that preserve the information of directions with 64 orientations and $S_5$ the finest scale, but has only one direction.

%Conforme a documentação presente em \cite{Candes2006}, o número total de coeficientes da decomposição em Curvelets e o número de escalas ($n_s$) depende do tamanho da imagem de entrada (M,N) e do número de direções ($\chi$) na segunda camada ($S_2$). Neste trabalho são utilizadas 32 direções na camada $S_2$ e todas as imagens são fragmentadas em blocos com tamanho (256,256). As camadas ainda não mencionadas são: $S_1$ que é camada mais grosseira e que possui apenas uma orientação, $S_3$ que possui 64 orientações, $S_4$ que é a camada mais fina que ainda preserva orientações, contando com 64 direções e a camada $S_5$ que é a mais fina de todas, porém não possui informação de orientação.

%, com sobreposição caso seja necessário. **Rever a necessidade da tabela**A Tabela \ref{tab:curvelet} mostra um resumo dos parâmetros das curvelets utilizados neste trabalho. 

%\begin{table}[H]
%		\renewcommand{\arraystretch}{1.3}
%		\caption{Estrutura dos coeficientes da Transformada Curvelet.}
%		\label{tab:curvelet}
%		\centering
%		\begin{tabular}{c c c c }
%			Característica & Camada ($S_j$)  & N. de coefic & N. Direções($\chi$)\\[1.5px] \hline
%			Grosseira & $S_1$  & 441 & 1  \\[1.5px] 
%			Detalhe & $S_2$   & 5984& 32\\[1.5px] 
%			Detalhe   & $S_3$  &  23232& 64\\[1.5px] 
%			Fina (direções)    & $S_4$  &  98624& 64\\[1.5px] 
%			Mais Fina  & $S_5$  & 65536& 1  
%		\end{tabular}
%	\end{table}
%**

The input image must be 8 bits. Color images or in other scales must be converted.
%Caso a imagem de entrada seja colorida ou em outra escala, deve ser feita a conversão para escala de cinza em 8 bits.
		
\subsection{Features extractions}
\label{ss:extract_features}

The model Curvelet2014\cite{Liu-2014} has 12 features grouped into 3 brands, Mean Energy by Scale (MES), Orientation Energy Distribution in $S_4$ layer (OED4) and Statistics of finest scale ($S_5$) (SFS5). This division of groups is maintained in this article.
	
%O modelo Curvelet2014\cite{Liu-2014} possui 12 característica agrupadas em 3 grupos, Energia média distribuída por Escala (EDE), Estatísticas da Energia por orientação na camada $S_4$(EOS4) e Estatísticas da camada $S_5$(ECS5). Essa divisão de grupos é mantida neste artigo.

Using the principal component analysis (PCA) in 12 original features of Curvelet2014, reveals 3 eigenvalues equal zero, this indicates that 3 out of 12 features are redundant. 

%Em todo caso, a utilização de uma transformação em componentes principais (PCA) revela que 3 das 12 características de Curvelet2014 são redundantes pois a transformação apresenta 3 autovalores iguais a zero.

\subsubsection{Mean Energy by Scale (MES)}

%As degradações que ocorrem nas imagens modificam seu espectro, essa modificação se manifesta inclusive no tamanho da janela da transformada Curvelet\cite{Liu-2014}.

The authors of \cite{Liu-2014} proposes the application of $\log_{10}$ on all the coefficients of the transform and the calculation of the mean per scale, as follows:

%Os autores de \cite{Liu-2014} propõe a aplicação de $\log_{10}$ sobre todos os coeficientes da transformada e o cálculo da média por escala, como se segue:
	
\begin{equation}
\bar{e}_{j}=E_g[\log_{10}(|S_j(\chi)|)], \quad j:[1,2,3,4,5] \;,\label{f:log10}
\end{equation}
	
\noindent where $E_g$ is the mean value of energy scale and $S_j(\chi_i)$ is the set of coefficients on scale $S_j$ with orientation $\chi$.
	
%\noindent em que $E_g$ denota o valor médio da energia por escala e $S_j(\chi_i)$ é o conjunto de coeficientes na escala $S_j$ com orientação $\chi$.

In \cite{Liu-2014},the authors make subtractions with energies  \eqref{f:log10} between the adjacent scales and interval scales. By inspection, it is possible to see that here live 2 of the 3 redundant features.

%Em \cite{Liu-2014}, os autores fazem subtrações com as energias \eqref{f:log10} entre as escalas adjacentes e saltando uma escala. Por inspeção, é possível identificar que parte das características redundantes denunciadas pelo PCA está presente neste grupo.

In other hand, the model M1 uses 3 combinations to discriminate the behavior of the coarser layers, $d_1=\bar{e}_1-\bar{e}_2$, $d_2=\bar{e}_2-\bar{e}_3$ e $d_3=\bar{e}_3-\bar{e}_4$. There is no redundancy between these features.

%Neste trabalho, o modelo M1 utiliza 3 combinações para discriminar o comportamento das camadas mais grosseiras, $d_1=\bar{e}_1-\bar{e}_2$, $d_2=\bar{e}_2-\bar{e}_3$ e $d_3=\bar{e}_3-\bar{e}_4$. Não há qualquer redundância entre essas características.
	
\subsubsection{Orientation Energy Distribution (OED4)}
\label{subsubsec:EOS4}

In mathematics, the anisotropy is the existence of a preferential direction. In images of natural scenes, textures and edges are sources of anisotropy\cite{Gabarda2007}.
	
%Em matemática, a anisotropia é a existência de uma direção preferencial. Nas imagens de cenas naturais, texturas e bordas são fontes de anisotropia\cite{Gabarda2007}.

In \cite{Gabarda2007} it is showed how the addition of white Gaussian noise (wn) and the Gaussian blurring process (gblur) impact on anisotropy measurements of the images. Also in \cite{Gabarda2007} it is said that artifacts created in compression like jpeg and jpeg2000 (jp2k) are difficult to handle.

%Em \cite{Gabarda2007} é mostrado como a adição de ruído branco Gaussiano (wn) e o processo de borramento Gaussiano (gblur) impactam em medidas de anisotropia das imagens. Também em \cite{Gabarda2007} é dito que artefatos criados em compactação como jpeg e jpeg2000 (jp2k) são de difícil tratamento.

In \cite{Sheikh} the authors argue the degradations affect the high-frequency components of the images. For model M1, the scale $S_4$ is the fine scale with different directions. The $ S_4 $ is the main source information for the study of anisotropy using the Curvelet transform.

%Para \cite{Sheikh}, degradações afetam mais as componentes de alta frequência das imagens, afirmando-se que as escalas mais finas são de crucial importância. Para o modelo M1, a camada $S_4$ é a mais fina que apresenta variação de direção, o que a torna $S_4$ a principal fonte informação para o estudo da anisotropia usando a transformada Curvelet.

There are more than 98 thousand coefficients in the $ S_4 $ scale, organized in a matrices and each matrix has 64 orientations ($ \chi $). The average energy per orientation in the layer $S_4$ can be measured as follows,

%Existem mais de 98 mil coeficientes na escala $S_4$, eles são organizados matricialmente e cada matriz tem 64 orientações ($\chi$). A energia média por orientação na camada $S_4$ pode ser medida da seguinte maneira,

\begin{equation}
EMO_{4}(\chi)=E[|S_{4}(\chi)|]\;,\label{f:energy_orient}
\end{equation}
	
\noindent where $E$ define the mean of coefficients in $S_{4}$ by direction and direction index $\chi$ has 64 values.

%\noindent onde $E$ define a média dos coeficientes de $S_{4}$ por direção e o índice de direção $\chi$ pode assumir 64 valores.

\begin{figure}[H]
\centering
\includegraphics[width=3.6in]{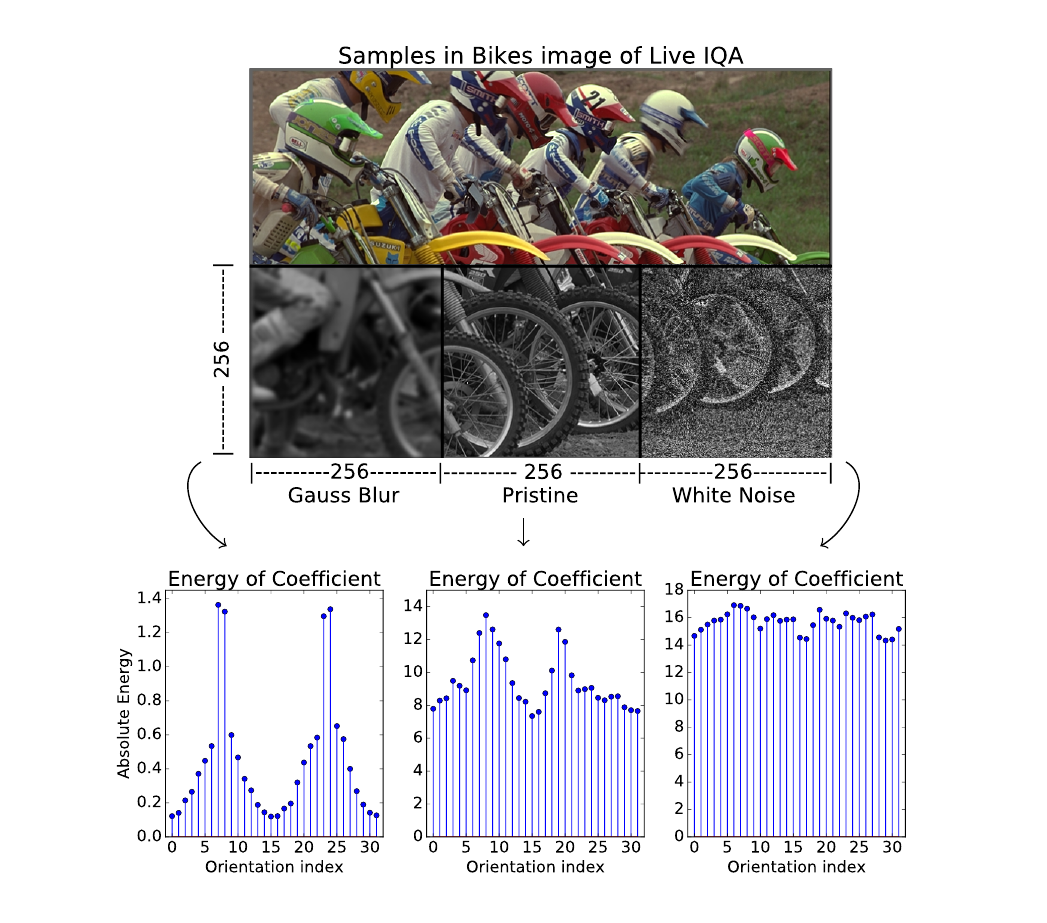}
		%%%%%%%%%%%%%%%%%%%%%%%%%%%%%%%%%%%%%%
%\caption{O quadro \textit{A} mostra uma representação artística dos blocos degradados sobrepostos a uma imagem do banco de dados LIVE IQA\cite{liveDatabase}. Quadro \textit{B} mostra a distribuição da médias dos coeficientes por orientação na camada $S_4$ de um bloco jpeg2000 (jp2k). O Quadro \textit{C}, mostra exemplos histogramas relacionados a camada $S_5$ para os quatro tipos de degradação abordados neste trabalho.}

\caption{The frame shows an artistic representation of the degraded blocks overlapping on an image of the LIVE IQA database\cite{liveDatabase} and respective $EMO_4$ vs $\chi$ graphs.}

\label{fig:bikes}
\end{figure}

The Figure \ref{fig:anisotropy}-A shows an image divided in 5 parts. The color part was preserved and the remaining parts were degraded. The Figure \ref{fig:anisotropy}-B shows the graphics $EMO_4$ vs $\chi$ for the jp2k part. In this image it is possible see the prominent points in the data. The Figure \ref{fig:bikes}  shows the behavior for classes Gaussian Blur (gblur), Gaussian white noise (wn) and the reference image.

%A Figura \ref{fig:anisotropy}-A apresenta uma imagem divida em 5 partes. A parte em cores foi preservada e as partes restantes sofreram degradações. A Figura \ref{fig:anisotropy}-B mostra o gráfico $EMO_4$ vs $\chi$ para um bloco jp2k. Nessa imagem é possível observar que existem pontos proeminentes nos dados. Esses pontos são \textit{outliers}, e eles se destacam nas imagens em que há borramento como uma anomalia\cite{Liu-2014}.

In \cite{Huber2009}, the authors argue that robust statistics means being callous to small deviations in assumptions. Since one of the sources of deviation is the outliers, then a strategy involving robust statistical descriptors is justified.

%Para \cite{Huber2009}, ser robusto significa ser insensível a pequenos desvios nos pressupostos. Como uma das fontes de desvio são os \textit{outliers}, então se justifica uma estratégia que envolva descritores estatísticos robustos.

Two complementary measures of dispersion are calculated in the scale $ S_4 $.

%Duas medidas complementares de dispersão são calculadas na escala $S_4$.

The first estimator is Quartile coefficient of dispersion (QCD)\cite{Bonett2006},

%A primeira baseada em Quartis, o Coeficiente de Variação entre Quartis (CVQ)\cite{Bonett2006},
	
\begin{equation}
CQV=\frac{Q_3-Q_1}{Q_3+Q_1}\quad,\label{f:coef_var_robust}
\end{equation}.

\noindent where $Q_1$ and $Q_3$ are first and third quartiles of data \eqref{f:energy_orient}.
	
%\noindent onde $Q_1$ e $Q_3$ são o primeiro e o terceiro quartis dos dados providos por (\ref{f:energy_orient}).

The second estimator is MAD\cite{Rousseeuw1993} divided by median, 

%A segunda é o estimador MAD\cite{Rousseeuw1993} dividido pela mediana, 

\begin{equation}
rMAD=\frac{med(|EMO_4(\chi)-Q_2|)}{Q_2}\quad,\label{f:rMAD}
\end{equation}

\noindent where $med$ computes the median, $Q_2$ is the median of the data \eqref{f:energy_orient}.

%\noindent onde $med$ denota cálculo da mediana, $Q_2$ corresponde a mediana dos dados providos por (\ref{f:energy_orient}).

The third feature in $S_4$ is the area below the curve of the data \eqref{f:energy_orient}.

%Uma terceira característica extraída em $S_4$ é a área abaixo da curva de delimitada pelos dados de (\ref{f:energy_orient}). 

All model-free features, applied to data directly.
	
\subsubsection{Statistics of finest scale $S_5$ (SFS5)} 

For Shen et al.\cite{Shen2009}, propose the transformation of the coefficients grouped into $S_5$,
	
%Em Shen et al.\cite{Shen2009}, as coordenadas do máximo de uma distribuição de probabilidades é baseada na seguinte transformação dos coeficientes agrupados em $S_5$,

\begin{equation}
e_{5}=\log_{10}(|S_5|)], \label{f:S5}
\end{equation}
	
\noindent where $S_5$ is the set of coefficients in scale 5. 

%\noindent onde $S_5$ é o conjunto de coeficientes da camada 5.

\begin{figure}[H]
\centering
\includegraphics[width=3.8in]{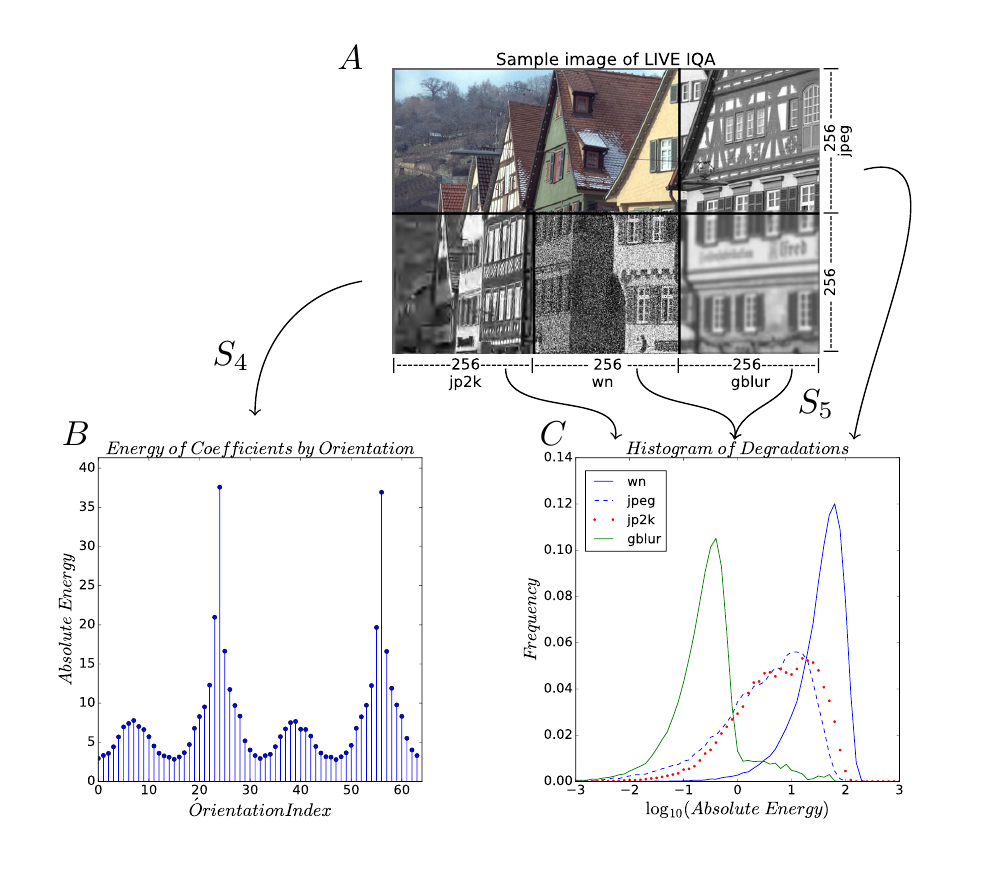}
		%%%%%%%%%%%%%%%%%%%%%%%%%%%%%%%%%%%%%%
%\caption{O quadro \textit{A} mostra uma representação artística dos blocos degradados sobrepostos a uma imagem do banco de dados LIVE IQA\cite{liveDatabase}. Quadro \textit{B} mostra a distribuição da médias dos coeficientes por orientação na camada $S_4$ de um bloco jpeg2000 (jp2k). O Quadro \textit{C}, mostra exemplos histogramas relacionados a camada $S_5$ para os quatro tipos de degradação abordados neste trabalho.}

\caption{The frame \textit{A} shows an artistic representation of the degraded blocks overlapping on an image of the LIVE IQA database\cite{liveDatabase}. The frame \textit{B} shows the mean distribution of coefficients by orientation in $S_4$ for the jpeg2000 (jp2k) block. The frame \textit{C} shows examples histograms related to layer $S_5$ for the four types of degradation used in this work.}

\label{fig:anisotropy}
\end{figure}

In Figure \ref{fig:anisotropy}-C shows the histogram of the probability distribution by type of degradation, what exposes several formats of distributions.

%A Figura \ref{fig:anisotropy}-C mostra exemplos de histogramas que representam as distribuições de probabilidade por tipo de degradação, na qual se observa grande variedade de formatos.

Several formats of distribution imply studying asymmetry and flattening of distributions. The lack of the normality condition and the presence of outliers justify doing this research with robust statistical descriptors\cite{Huber2009}.

%Os diversos formatos das distribuições leva a investigar as curvas utilizando momentos estatísticos relacionados a assimetria e achatamento. A ausência da condição de normalidade e a presença de \textit{outliers} justificam fazer essa investigação com descritores estatísticos robustos\cite{Huber2009}.

The robust descriptors in SFS5 are better understood using the octiles concept, which divides the sample into 8 sections of equal size, $Oc_1$ to $Oc_7$. This method is applied directly to the coefficients transformed by \eqref{f:S5} without requiring a histogram or curve fit.

%Os descritores robustos para ECS5 são melhor entendidos utilizando o conceito de octis, que divide a amostra em 8 seções de igual tamanho, $Oc_1$ até $Oc_7$. Esse método é aplicado diretamente aos coeficientes transformados por (\ref{f:S5}) sem a necessidade de recorrer a histograma ou ajuste de curvas. 

The first three descriptors are the median $(Oc_4)$, interquartile range ($Oc_6-Oc_2$) and the MAD estimator $(med(e_5-Oc_4))$.

%Os três primeiros descritores são a mediana ($Oc_4$), a distância entre quartis ($Oc_6-Oc_2$) e do estimador MAD ($med(e_5-Oc_4)$).

The fourth descriptor is the Bowley Skewness\cite{Groeneveld1984}, 
%O quarto descritor é a assimetria de Bowley\cite{Groeneveld1984}, 

\begin{equation}
\gamma = \frac{Oc_6+Oc_2-2Oc_4}{Oc_6-Oc_2}\quad,\label{f:skew_bow}
\end{equation}

\noindent the fifth feature is Moors Kurtosis\cite{Moors1988},  

%\noindent o quinto é a curtose de Moors\cite{Moors1988}, 

\begin{equation}
\kappa= \frac{(Oc_7-Oc_5)+(Oc_3-Oc_1)}{Oc_6-Oc_2}\quad,\label{f:kurt_moors}
\end{equation}

\noindent where the octiles are define with percentiles, $Oc_1=p(12.5\%)$, $Oc_2=p(25\%)$, $Oc_3=p(37.5\%)$, $Oc_4=p(50\%)$, $Oc_5=p(62.5\%)$, $Oc_6=p(75\%)$, $Oc_7=p(87.5\%)$, besides $Oc_{2i}=Q_i$ for $i=\{1,2,3\}$.

%\noindent onde os octis são definidos com base nos percentis, sendo $Oc_1=p(12.5\%)$, $Oc_2=p(25\%)$, $Oc_3=p(37.5\%)$, $Oc_4=p(50\%)$, $Oc_5=p(62.5\%)$, $Oc_6=p(75\%)$, $Oc_7=p(87.5\%)$, note que $Oc_{2i}=Q_i$ para $i=\{1,2,3\}$.

\subsection{Post Processing}

Remembering, the model M1 has 11 features without redundancy, 3 in MES, 3 in OED4 and 5 in SFS5, calculated for each $256\times 256$ gray scale block.

%O modelo proposto neste artigo (M1), possui 11 características não redundantes, 3 oriundas de EDE, 3 de EOS4 e 5 de ECS5, que são obtidas para cada bloco de tamanho $256\times 256$ em escala de cinza.

For each input image, the number of blocks varies according its size. The 11 features are calculate for each block. The simple mean of each features make the 11-dimension vector. This vector represent the input image on features space and is the input of 2-stages SVM for machine learnig.

%Para cada imagem de entrada, o número de blocos varia conforme seu tamanho original e as 11 características são calculadas para cada bloco. A média aritmética de cada característica forma um vetor 11-dimensional que representa a imagem de entrada no espaço de características. Esse vetor é enviado como entrada para a estrutura computacional em 2-\textit{stages} SVM para o aprendizado de máquina.

\subsection{Architecture 2-stages SVM}
\label{ss:2stages}

In this work, the 2-stages SVM uses the 11-dimension vector obtained by the coefficients of the Curvelet transform. This vector passes through ensemble of regressors. The predictions of regressors are fused using mean weight. The weights are provided by the classifier. 

%Nesse trabalho, a estrutura em 2 estágios, tem como entrada principal o vetor 11-dimensional extraído dos coeficientes da transformada Curvelet. Esse vetor é processado por um banco de regressores, cujas predições são fundidas usando uma média ponderada em que os pesos são dados pelo classificador.

In Figure \ref{fig_sim}, 4 branches in dashed box show the set of regressors. Each regressor uses an $\nu$-SVR with Radial Base Funcition (RBF) kernel. Each regressor was trained with degraded images of a specific class, becoming a specialist in that class.

%Na Figura \ref{fig_sim}, o ramo SVR do diagrama mostra o conjunto de regressores. Cada regressor usa uma arquitetura $\nu$-SVR com núcleo do tipo função de base radial (RBF). Cada regressor é treinado com imagens degradadas de uma classe específica, tornando-se especialista naquela classe.

For this work, the set of regressors returns a column vector $ 4 \times 1 $ $( \vec{q} )$ with the predictions of each regressor.

%Para esse trabalho, o banco de regressores retorna um vetor coluna $4\times 1$ ($\vec{q}$) com as predições de cada regressor.

Also in Figure \ref{fig_sim}, the upper branch is the classifier. We used a C-SVC with output in probability and an RBF kernel. The output of classificator is a column vector  $4\times 1$ $(\vec{p})$.

%Ainda na Figura \ref{fig_sim}, o ramo superior é o classificador, foi usado um C-SVC com saídas em probabilidade e um núcleo RBF. A saída do classificador é um vetor coluna $4\times 1$ ($\vec{p}$).

Each module can be approached independently. In this work, the only additional procedure was the application of a standardized scale transformation (zero mean and unit variance) for each block.

%Cada módulo pode ser abordado de forma independente a partir daqui. Neste trabalho, o único procedimento adicional foi a aplicação de uma transformação para escala padronizada (média zero e variância unitária) das características para cada bloco.

When a test image is inserted in this architecture, it passes through the classifier and the 4 regressors. The objective quality index $(Q)$ is the weighted average of the predictions of the regressors bank, expressed by,
	
%Quando uma imagem de teste é inserida nessa arquitetura, ela passa pelo classificador e pelos 4 regressores. O índice de qualidade objetivo ($Q$) é a média ponderada das predições do banco de regressores, expressa por,
	
\begin{equation}
Q=\vec{p}^{\;T}\;\vec{q}\;.\label{f:Qpq}
\end{equation}

\section{Methodology}
\label{s:method}
	
\subsection{Data set of images}

\subsubsection{LIVE IQA}

The set LIVE IQA\cite{liveDatabase}\footnote{\url{http://live.ece.utexas.edu/research/quality/subjective.htm}} has 29 reference images of natural scenes. From them they create 779 degraded images, distribute in 5 categories. The subjective quality index used in the LIVE IQA survey is the Differential Mean Opinion Scores (DMOS) with values between 0 and 100, where 0 values indicates the highest quality.

%O banco de dados LIVE IQA\cite{liveDatabase} constitui-se de 29 imagens de referência de cenas naturais, que originam 779 imagens degradadas em 5 categorias. O índice de qualidade subjetivo utilizado na pesquisa do LIVE IQA é o \textit{differential mean opinion scores} (DMOS) com valores de 0 até 100.

In this work we will use only 4 classes: jpeg (jpeg) compression and jpeg2000 (jp2k), white noise Gaussian (wn), Gaussian blur (gblur), total number of 634 degraded images.

%Nesse trabalho usaremos apenas 4 categorias: compressão jpeg (jpeg) e jpeg2000 (jp2k),  ruído branco Gaussino (wn), borramento Gaussiano (gblur), número total de 634 imagens degradadas.

\subsubsection{TID2013}
	
The set TID2013\cite{TID2013}\footnote{\url{http://www.ponomarenko.info/tid2013/tid2013.rar}} has 25 reference images, 24 of natural scenes and one artificial. It has 25 classes of distortion with 5 levels of degradation per class, having a total of 3125 degraded images with standard size $512\times 384$. The subjective quality index is the Mean Opinion Scores (DMOS) with values between 0 and 9, where 9 values indicates the highest quality.

%O banco de dados TID2013\cite{TID2013} possui 25 imagens de referência, 24 naturais e uma artificial. Dispõe-se de 25 categorias de distorção e 5 níveis de degradação por categoria, tendo um total de 3125 imagens degradadas de tamanho padronizado $512\times 384$. O índice de qualidade subjetivo utilizado é o \textit{mean opinion scores} (MOS), com faixa de valores entre 0 e 9.

For the experiments of this work only the natural images and the 4 classes used in LIVE IQA . The total of the images used is 480.

%Para os experimentos deste trabalho são utilizadas apenas as imagens naturais e as 4 classes utilizadas no LIVE IQA. O total de imagens utilizadas é de 480.

\subsubsection{CSIQ}

The set CSIQ\cite{CSIQ}\footnote{\url{http://vision.eng.shizuoka.ac.jp/mod/page/view.php?id=23}} has 30 reference images. There are 6 classes of distortion and 5 levels of degradation per class, generating a total of 900 degraded images of standard size $ 512 \times 512 $. The subjective quality index used in the CSIQ survey is the Differential Mean Opinion Scores (DMOS) with values between 0 and 1, where 0 values indicates the highest quality.

%O banco de dados CSIQ\cite{CSIQ} possui 30 imagens de referência. São 6 classes de distorção e 5 níveis de degradação por classe, gerando um total de 900 imagens degradada de tamanho padronizado $512\times 512$. O índice de qualidade subjetivo utilizado na pesquisa do CSIQ é o \textit{differential mean opinion scores} (DMOS) com valores de 0 até 1.

In the experiments, four previously mentioned categories were used. The total of 600 degraded images.

%Nos experimentos são utilizadas 4 categorias mencionadas anteriormente, num total de 600 imagens degradadas.	

The data set LIVE IQA and TID2013 has the same primary font for reference images, a set of public images Kodak Lossless True Color Image Suite\footnote{\url{http://r0k.us/graphics/kodak/}}\cite{TID2013}. To perform an experiment absolutely independent of any bias resulting from this content overlap we use the CSIQ data set.

%Os bancos de dados LIVE IQA e TID2013 possuem a mesma fonte para imagens de referência, um conjunto de imagens públicas da Kodak\cite{TID2013}. Para realizar um experimento absolutamente independente de qualquer viés fruto dessa sobreposição de conteúdo buscou-se a base CSIQ que utiliza outras imagens de referência.

\subsection{Machine Learning}

The following steps are applied to model M1 and concurrent job Curvelet2014 \cite{Liu-2014}.

%Para possibilitar a comparação entre os modelos concorrentes, a lista de procedimentos a seguir é aplicada ao modelo M1 e ao trabalho concorrente Curvelet2014\cite{Liu-2014}.

In this work, the training is performed with the degraded images of LIVE IQA.

By \cite{Moorthy2010} and \cite{Liu-2014}, the training set should be separated using the reference images. This prevents content overlap in the test involving LIVE IQA.

%Conforme \cite{Moorthy2010} e \cite{Liu-2014}, o conjunto de treino deve ser separado usando as imagens de referência para evitar sobreposição de conteúdo no teste envolvendo LIVE IQA.

The separation of degraded images is done by mapping reference images. We use 5-fold method with 40 repetitions to cover a large number of possibilities of association of reference images. This makes a 200 different sets to train. The training is done with the degraded images mapped using 80\% of the reference images in each set. The rest of the degraded images, are used in the test phasis.

%Nesse trabalho, o treino é realizado com as imagens degradadas do LIVE IQA. A separação das imagens degradadas é feita mapeando imagens de referencia. Para cobrir um grande número de possibilidades de associação das imagens de referência, é usado o método de 5-\textit{fold} com 40 repetições distintas e aleatórias, que resulta em um total de 200 configurações distintas. O treino é feito com as imagens degradadas mapeadas usando 80\% das imagens de referência em cada configuração. O restante das imagens degradadas é direcionado para teste.

After training, the models are tested separately using 20 \% of the degraded images of the LIVE IQA, 480 images of the TID2013 set and 600 images of the CSIQ set.

%Após treinado, os modelos são testados separadamente usando 20\% das imagens degradadas do LIVE IQA, 480 imagens do conjunto TID2013 e 600 imagens do conjunto CSIQ.

The SVM hyperparameters must be obtained for each training set\cite {Moorthy2010}.

%Segundo Moorthy e Bovik\cite{Moorthy2010} os hiperparâmetros do SVM devem ser obtidos para cada conjunto de treino. 

For each round the hyperparameters $(C, \gamma)$ are computed for C-SVC using the stratified 5-fold cross-validation method with 5 random repetitions. As same, it is computed to each regressor $\nu$-SVR the hyperparameters  $(C,\gamma,\nu)$ using 5-fold cross-validation method with 5 random repetitions. The search grids for each of the hyperparameters were $C:[2^{-1},2^1,2^3,2^5,2^7,2^9,2^{11},2^{13}]$ , $\gamma:[2^{-8},2^{-6},2^{-4},2^{-2},1]$. The parameter $\nu$ was set at 0.5 because tests using a search grid between 0.25 and 0.75 showed minimal variation of results.

%Para cada rodada é computado para C-SVC os hiperparâmetros (C,$\gamma$) usando o método de validação cruzada 5-\textit{fold} estratificado com 5 repetições aleatórias. Da mesma maneira, é computado para cada regressor $\nu$-SVR os hiperparâmetros (C,$\gamma$,$\nu$) usando o método de validação cruzada 5-\textit{fold} com 5 repetições. As grades de busca de cada um dos hiperparâmetros foram $C:[2^{-1},2^1,2^3,2^5,2^7,2^9,2^{11},2^{13}]$ , $\gamma:[2^{-8},2^{-6},2^{-4},2^{-2},1]$. O parâmetro $\nu$ foi fixo em 0,5 pois testes utilizando uma grade de busca entre os valores 0,25 e 0,75 apresentaram variações mínimas de resultado.

This computational implementation was performed using the python scikit-learn 0.19.1 package\footnote{\url{https://scikit-learn.org/stable/}}.

%Esta implementação computacional foi realizada utilizando o pacote python scikit-learn 0.19.1.
	
\subsection{Quality Measures of Predictions}
\label{ss:qualidade}

The main product of NR methods is the objective quality index(Q). The quality of the prediction is measured using correlation between Q index of the images and the subjective index associated of the images (DMOS or MOS).

%O principal produto dos métodos NR é o índice de qualidade objetivo (Q), e a qualidade dessa predição é mensurada usando medidas de correlação entre índice Q e o índice subjetivo associado a imagem degradada (DMOS ou MOS).

Ponomarenko et al.\cite{TID2013} recommend the Spearman's rho (SROCC) and the Kendall's tau (KROCC) as correlation measures. Values close to 1 indicate a better correlation with human perception.

%Nesse trabalho são usados os coeficiente de correlação de postos de Spearman (SROCC) e o  coeficiente de correlação de postos de Kendall (KROCC) pois essas são as correlações recomendadas na referência\cite{TID2013}. Valores próximos de 1 indicam melhor correlação com a percepção humana.

To measure the quality of the classifier the accuracy is used. Values closer to 1 indicate a better classification.	

%Medir a qualidade do classificador é relevante, e essa medida é feita usando-se a acurácia. Valores mais próximos de 1 indicam melhor classificação.

\subsubsection{Statistical significance}

Statistical significance: Due to the structural proximity
between the competing models, we chose a statistical
significance test to support more accurate conclusions. And, when
normality and homoscedasticity tests indicate that there is
no distribution common to all experiments non-parametric
test is recommended.

%Neste trabalho, para se ter resultado mais acurados, tendo em vista a proximidade estrutural entre os modelos concorrentes optou-se por recorrer a teste de significância estatística.

%Os testes de normalidade e homoscedasticidade indicam que não há uma distribuição comum a todos os experimentos. Isso indica a necessidade de um teste de significância não paramétrico. 

The way the experiments were conducted generates paired results. The Paired Samples Wilcoxon Test\cite{Wilcoxon1945} is adequate and was used in the measures of SROCC, KROCC and accuracy of the M1 and Curvelet2014.

%Por sua vez a maneira como os experimentos foi conduzida gera resultados pareados, portanto o teste de postos pareados de Wilcoxon\cite{Wilcoxon1945} é adequado e foi utilizado nas medidas de SROCC, KROCC e acurácia dos modelos M1 e Curvelet2014.

For this test the null hypothesis is: The means of the groups compared is identical. The alternative hypothesis is: The means are different.

%Para esse teste a hipótese nula é que as médias dos grupos comparados é idêntica. A hipótese alternativa é de que as médias são diferentes.

%Nesse artigo a hipótese nula é rejeitada para valor $p<0.05$, as comparações em que isso ocorrer a média de maior valor é destacada.

In this article the null hypothesis is rejected for value $ p <0.05 $, the comparisons in which this occurs the mean of greater value is highlighted. There is greater interest in cases where the null hypothesis is rejected. It is known that the rejection of the null hypothesis does not imply accepting the alternative hypothesis.

The Wilcoxon test will only allow a simple and reasonable conjecture that there is evidence to believe that the highest average model has a better result.

%Neste artigo há maior interesse nos casos em que a hipótese nula é rejeitada. É sabido que a rejeição da hipótese nula não implica em aceitar a hipótese alternativa. O teste de Wilcoxon pareado permite no máximo a conjectura simples e razoável de que existem indícios para acreditar que o modelo com maior média possui um resultado melhor. 

To strengthen this conjecture, we will emphasize results that are persistent in at least 2 tests and that there is no setback in the remaining tests.

%Visando robustecer essa conjectura, serão enfatizados resultados que sejam persistentes em pelo menos 2 testes e que nos testes restantes pelo menos seja aceita a hipótese nula.

This computational implementation is performed in subsection \ref{ss:qualidade} were made using the scipy 1.1.0 python package\footnote{\url{https://www.scipy.org/}}.

%Esta implementação computacional e a realizada na subseção \ref{ss:qualidade} foram feitas utilizando o pacote python scipy 1.1.0.

\section{Experiments and Results}
\label{s:exp_res}

We made two types of experiments. The first evaluating the overall result using the four classes of degradation simultaneously and the second evaluating the result by class.

%Foram realizados dois tipos de experimentos, o primeiro avaliando o resultado geral utilizando as 4 classes de degradação simultaneamente e o segundo avalia o resultado por classe.

The cells of Tables \ref{res:res_geral} and \ref{res:res_class} should be evaluated in pairs (white and gray). There is more emphasis in the analysis by columns, they go through the 3 tests. Do not compare the values of SROCC and KROCC because they evaluate different forms of correlation.

%As células das Tabelas \ref{res:res_geral} e \ref{res:res_class} devem ser avaliadas aos pares(branca e cinza). Há maior enfase na análise por colunas, elas percorrem os 3 testes. Não se deve comparar os valores de SROCC e KROCC eles avaliam formas distintas de correlação.

\subsection{Four Classes test}

This experiment consists of testing competing models with all test images. The outputs $ Q $ and the output of the classifier are collected. Predictions are compared to subjective quality index and degraded image label.

%Esse experimento consiste em testar os modelos concorrentes com todas as imagens de teste sem restrições. São recolhidas as saídas $Q$ e a saída dos módulo classificador. Essas saídas preditas são comparadas o índice de qualidade subjetivo e com as etiqueta da imagem degradada.

The same sets of images are used for training and testing in both models over 200 rounds.

%São utilizados rigorosamente os mesmos conjuntos de imagens para treino e teste em ambos os modelos ao longo de 200 rodada.

\label{ss:Multiclass}
\begin{table}[H]
		\centering
		\renewcommand{\arraystretch}{1.3}
		\caption{Mean values of SROCC, KROCC and Accuracy values in 200 rounds, models trained with 80 \% of LIVE IQA and test indicated in the table. White lines for model M1 and gray lines for model Curvelet2014\cite{Liu-2014}. Values in bold are indicative of improvement in which the null hypothesis was rejected.
		}
		%\caption{Médias dos valores de SROCC, KROCC e Acurácia em 200 realizações distintas, modelos treinados com 80\% do LIVE IQA e teste indicado na tabela. Linhas brancas para o modelo M1 e linhas cinza para modelo Curvelet2014\cite{Liu-2014}. Valores em negrito são indicativos de melhoria em que a hipótese nula foi rejeitada.}
		\label{res:res_geral}
		\begin{tabular}{c|ccc}
			&SROCC &KROCC &Acurácia \\\hline
			\multicolumn{4}{c}{LIVE IQA 20\%}  \\[0.25px] \hline 	
M1 &0.8795&\textbf{0.7392}&\textbf{0.8627}\\[0.25px]
			\cellcolor{gray!25}Curvelet2014\cite{Liu-2014}&\cellcolor{gray!25}0.8762&\cellcolor{gray!25}0.7198 &\cellcolor{gray!25}0.8248 \\[0.25px]
			\multicolumn{4}{c}{TID2013} \\[0.25px] \hline 
			M1 &0.8392&0,6541&\textbf{0.8148}\\[0.25px]
			\cellcolor{gray!25}Curvelet2014\cite{Liu-2014}& \cellcolor{gray!25}0.8401 &\cellcolor{gray!25}0.6525&\cellcolor{gray!25}0.7791\\[0.25px]
			\multicolumn{4}{c}{CSIQ} \\ \hline 
			M1 &\textbf{0.7764}&\textbf{0.5834}&\textbf{0.7235}\\[0.25px]
			\cellcolor{gray!25}Curvelet2014\cite{Liu-2014}& \cellcolor{gray!25}0.7470 &\cellcolor{gray!25}0.5470&\cellcolor{gray!25}0.6905
		\end{tabular}
	\end{table}

The analysis of Table \ref{res:res_geral} column by column shows that the M1 classify demonstrates better than Curvelet2014. 

Also in Table \ref{res:res_geral} SROCC indicates that in 2 tests the null hypothesis is verified. On the other hand KROCC presents 2 results in which it is rejected, both favorable to the model M1.

%A análise da Tabela \ref{res:res_geral} coluna a coluna, possibilita concluir que o modelo M1 classifica melhor que Curvelet2014. Essa mesma tabela SROCC indica que em 2 teste a hipótese nula se verifica. Por outro lado KROCC apresenta 2 resultados em que ela é rejeitada, ambos favoráveis ao modelo M1.

%%Não presente no trabalho% Em cada coluna ao se subtrair o maior e menor valor, obtemos a variância atribuída a cada modelo. Para SROCC M1  apresenta variâncias de 10.3\% e Curvelet2014 12.9\%. Para KROCC M1 tem 15.6\% e Curvelet2014 17.3\%. Para acurácia M1 apresenta 13.9\% e Curvelet2014 13.4\%. Esse resultado indica que M1 apresenta uma variância menor de seus resultados, algo desejado dentro desse tipo de trabalho.%Não presente no trabalho%

Separating table \ref{res:res_geral} by test, M1 shows better results in the LIVE IQA 20\% and CSIQ tests. The test performed with TID2013 shows statistically equivalent results.

%Analisando cada teste da Tabela \ref{res:res_geral} separadamente, constata-se que o modelo M1 apresenta resultados melhores nos testes LIVE IQA 20\% e no teste CSIQ. O teste realizado com TID2013 apresenta resultado estatisticamente equivalentes.

In this experiment there were 6 favorable results for M1, 3 indifferent results for the M1 model and none was unfavorable.

%Neste experimento, houveram 6 resultados favoráveis a M1, 3 resultados indiferentes ao modelo M1 e nenhum resultado desfavorável.
	
\subsection{Performance by class of degradation}
\label{ss:single_class}

This experiment discriminates which degradations are best described by each model. To do this, the $ Q $ prediction is made using only images of a particular degradation class, and the accuracy is not evaluated.

%Visando discriminar quais degradações são melhor descritas por cada modelo, a predição $Q$ é feita utilizando apenas imagens de uma determinada classe de degradação, e a acurácia não é avaliada.

Four different experiments are performed, one for each class of degradation. The results are grouped in the Table  \ref{res:res_class}.

%Apesar de ser colocado dentro de uma única tabela, são realizados 4 experimentos distintos, um para cada classe de degradação.

The analysis of Table \ref{res:res_class} column by column shows that M1 has superior performance in the jp2k and wn classes. For jpeg class KROCC gave indications that Curvelet2014 is better to describes this class, but this conclusion is contradicted by the SROCC test of TID2013. The gblur class does not present results that are favorable to any model.

%Analisando a Tabela \ref{res:res_class} coluna a coluna, é possível ver que M1 tem destaque na classe jp2k e na classe wn. Na classe jpeg KROCC deu indícios de que Curvelet2014 descreva melhor essa classe. Esse resultado é contradito pelo teste SROCC de TID2013. O classe gblur não apresenta resultados que sejam favoráveis a qualquer dos modelos.

There was no systematic behavior for the averages of the gblur class. It is possible to make a variation analysis, where smaller variations are preferable to larger ones. 

In this analysis for class gblur, it is verified that M1 is superior to the competitor. The model M1 has variations 12.1\% for SROCC and 27.8\% for KROCC. The model Curvelet2014 has variatons 20.6\% for SROCC and 35.7\% for KROCC.

%Apesar de não haver um comportamento sistemático para as médias da classe gblur, é possível recorrer a uma análise de variação entre os 3 testes. Variações menores são preferíveis às maiores. Nessa análise é verificado que M1 possui variação de SROCC em 12,1\% e Curvelet2014 em 20,6\% e variação de KROCC para M1 27,8\% e Curvelet2014 35,7\%.

\begin{table}[H]
	\setlength{\tabcolsep}{4pt}
	\renewcommand{\arraystretch}{1.3}
	\caption{Mean values of SROCC, KROCC by type of degradation in 200 rounds, models trained with 80 \% of LIVE IQA and test indicated in the table. White lines for model M1 and gray lines for model Curvelet2014\cite{Liu-2014}. Values in bold are indicative of improvement in which the null hypothesis was rejected.	}
	%\caption{Médias dos valores de SROCC, KROCC por tipo de degradação em 200 realizações distintas, modelos treinados com 80\% do LIVE IQA e teste indicado na tabela. Linhas brancas para o modelo M1 e linhas cinza para modelo Curvelet2014\cite{Liu-2014}. Valores em negrito são indicativos de melhoria em que a hipótese nula foi rejeitada.}
\label{res:res_class}
	\centering
		\begin{tabular}{c|c|c c c c} 
			& &  jp2k & jpeg & wn & gblu\\[0.5px] \hline
            \multicolumn{6}{c}{LIVE IQA 20\%}  \\[0.25px] \hline 
	\multirow{2}{*}{SROCC}	& M1 &\textbf{0,8743}&0,8805& 1 &0,9045\\[0.25px] 
	& Curvelt2014\cite{Liu-2014}\cellcolor{gray!25} &0,8373\cellcolor{gray!25} &\cellcolor{gray!25}\textbf{0.9309}&\cellcolor{gray!25}0,9990&0,9175\cellcolor{gray!25}\\[0.25px]
 \multirow{2}{*}{KROCC} &  M1 &\textbf{0,7867} &0,8116&1 &0,8850\\ [0.25px]
	& \cellcolor{gray!25}Curvelet2014\cite{Liu-2014} &0,7337\cellcolor{gray!25} &\cellcolor{gray!25}\textbf{0,8676}&\cellcolor{gray!25}0,9980 &0,8830\cellcolor{gray!25}\\ [0.25px]
            \multicolumn{5}{c}{TID2013}  \\[0.25px] \hline 
	\multirow{2}{*}{SROCC}	& M1 &\textbf{0,8447}&\textbf{0,8365} &\textbf{0,9058}&0,8548\\[0.25px] 
	&Curvelt2014\cite{Liu-2014}\cellcolor{gray!25}&0.7703\cellcolor{gray!25} &\cellcolor{gray!25}0,8337&\cellcolor{gray!25}0,8784&\textbf{0,8608}\cellcolor{gray!25}\\[0.25px]
 \multirow{2}{*}{KROCC} &  M1 &\textbf{0,6467}&0,6337&\textbf{0.7284} &0,6592\\ [0.25px]
	&\cellcolor{gray!25}Curvelet2014\cite{Liu-2014} &0,5742\cellcolor{gray!25} &\cellcolor{gray!25}0,6353&\cellcolor{gray!25}0,6967&\textbf{0,6658}\cellcolor{gray!25}\\ [0.25px]
     
  \multicolumn{5}{c}{CSIQ} \\ \hline 
	\multirow{2}{*}{SROCC}	& M1 &\textbf{0,7964}&0,7747&\textbf{0,8875}&\textbf{0,7836}\\[0.25px] 
	&Curvelt2014\cite{Liu-2014}\cellcolor{gray!25}&0,7252\cellcolor{gray!25} &\cellcolor{gray!25}\textbf{0,7861}&\cellcolor{gray!25}0,8674&\cellcolor{gray!25}0,7105\\[0.25px]
 \multirow{2}{*}{KROCC} &  M1 &\textbf{0,5868}&0,5670&\textbf{0,6989} &\textbf{0,6070}\\ [0.25px]
	& \cellcolor{gray!25}Curvelet2014\cite{Liu-2014} &0,5230\cellcolor{gray!25} &\cellcolor{gray!25}\textbf{0,5750}&\cellcolor{gray!25}0,6773&\cellcolor{gray!25}0,5255
		\end{tabular}
	\end{table}

%A variância apresentada por classe em cada modelo foi SROCC M1 7.8\%, 10.6\%, 11.2\% e 12\% e SROCC Curvelet2014 11.2\%, 14.5\%, 13.2\% e 20.6\%, ambos na ordem que segue a tabela. Da mesma forma para KROCC M1 20.0\%, 24.5\%, 30.1\% e 27.8\% e KROCC Curvelet2014 21.1\%, 29.3\%, 32.1\% e 35.7\%. Esses resultado indicam que para todas as classes M1 apresenta menor valor de variância.

In this experiment there were 13 favorable results for M1, 2 indifferent results for the M1 model and 6 unfavorable results.

\section{Conclusion}
\label{s:Conclusion}

In this work, an objective method was developed to evaluate image quality without reference image, acronym (NR IQA). The method consists in extracting 11 features within the transformed space of the Curvelets and has as a differential the use of robust statistical descriptors. The proposed model M1 positions itself as an alternative to the model proposed in \cite{Liu-2014}, Curvelet2014. The models were tested with 3 databases of natural scenes images.

%Nesse trabalho foi desenvolvido um método objetivo para avaliação de qualidade de imagens sem a imagem referência, sigla (NR IQA). O método consiste em extrair 11 características dentro do espaço transformado das Curvelets e tem como diferencial o uso de descritores estatísticos robustos. O modelo proposto M1 se posiciona como uma alternativa ao modelo proposto em \cite{Liu-2014}, Curvelet2014. Os modelos foram testados com 3 bancos de dados de imagens de cenas naturais. 

The M1 model classify better than Curvelet2014 according to Table \ref{res:res_geral}. The probable hypothesis is that the features of M1 are more discriminatory for classification, there were already indications of this in the transformation of principal components for the 12 characteristics of Curvelet2014. 

Also, Table \ref{res:res_geral} shows improvements in agreement between Q and subjective indexes (DMOS and MOS) in a multi-class experiment. In such experiment, 6 of them give favorable results for M1, 3 of them give indifferent results for the M1 model and none was unfavorable.

%O modelo M1 classifica melhor que Curvelet2014 conforme a Tabela \ref{res:res_geral}. A hipótese provável é que as características de M1 sejam mais discriminativas para classificação, já haviam indícios disso na transformação de componentes principais para as 12 características de Curvelet2014.

The M1 model also presents the highest correlation for the jp2k class for all tests and correlations. The most likely hypothesis is that robust statistical descriptors will better discriminate the data from the jp2k class. This class is quite problematic, as shown in Figure \ref{fig:anisotropy} this class can present outliers in $S_4$ and has a no normal distribution of coefficients in $S_5$.

%O modelo M1 também apresenta maior correlação para a classe jp2k para todos os testes e correlações. A hipótese mais provável é de que os descritores estatísticos robustos discriminem melhor os dados da classe jp2k. Como foi mostrado na Figura \ref{fig:anisotropy} essa classe pode apresentar \textit{outliers} em $S_4$ e a distribuição em $S_5$ geralmente foge da condição de normalidade.

The class jpeg presents a KROCC measure in favor of the model Curvelet2014. This indication is not strong, because is contradicted by a SROCC test.

%A classe jpeg apresenta para medida KROCC um favorecimento do modelo Curvelet2014. Por ser contradito por um teste SROCC essa indicação não é forte.

The M1 model presents a better correlation for the degradation of the white Gaussian noise type (wn) in most tests. There was no evidence of the contrary.

%O modelo M1 apresenta melhor correlação para a degradação do tipo ruído branco Gaussiano (wn) na maioria dos testes, e não houve qualquer indício contra sua superioridade frente ao modelo Curvelet2014.

For Gaussian blurring degradation (gblur) the M1 model presented more robust and unbias results. It is callous to bias introduced by training in a specific database. This last statement is also endorsed by superior performance in the CSIQ database.

%Para a degradação borramento Gaussiano (gblur) o modelo M1 apresentou resultado mais robusto e resistentes a viés introduzidos pelo treinamento em um banco de dados específico. Essa última afirmação também é endossada pelo desempenho superior no banco de dados CSIQ.

The statistical tests performed showed that, more often, M1 model presented superior results. M1 presented higher correlation with statistically significant in almost all the tests and with smaller numerical variation among the databases. 

For these reasons the M1 model is the most advantageous option among the models tested.

%Os teste estatísticos realizados mostraram que modelo M1 apresentou resultados superiores,  estatisticamente significantes em quase todos os testes e com menor variação numérica entre os bancos de dados. Por esses motivos o modelo M1 é a opção mais vantajosa entre os modelos testados.

	
	
	%\bibliographystyle{IEEEtran}
	\bibliography{./MyCollection.bib}
	%\begin{thebibliography}{1}
	
	%\end{thebibliography}

\end{document}